\pdfoutput=1

\documentclass[11pt]{article}

\usepackage[preprint]{acl}
\usepackage{booktabs}
\usepackage{csvsimple}
\usepackage[utf8]{inputenc}
\usepackage{multirow}
\usepackage{array}
\usepackage{makecell}
\usepackage{times}
\usepackage{latexsym}

\usepackage{xcolor}

\usepackage[T1]{fontenc}

\usepackage[utf8]{inputenc}

\usepackage{microtype}

\usepackage{inconsolata}

\usepackage{graphicx}

\usepackage{ifthen} 
\usepackage{url} 
\usepackage{acronym}
\usepackage[inline]{enumitem}

\acrodef{LLM-MAS}{LLM-based Multi-Agent Systems}
\acrodef{MAS}{Multi-Agent Systems}
\acrodef{LLM}{large language model}
\acrodef{RL}{reinforcement learning}
\acrodef{NLP}{natural language processing }

\title{A Survey on LLM-based Multi-Agent System: \\ Recent Advances and New Frontiers in Application}

\author{
  Shuaihang Chen$^1$ \quad
  Yuanxing Liu$^1$ \quad
  Wei Han$^1$ \quad
  Weinan Zhang$^1$\footnotemark[2] \quad
  Ting Liu$^1$  \quad \\
  $^1$\normalsize{Research Center for Social Computing and Information Retrieval}\\[-.05cm]
  \normalsize{Harbin Institute of Technology, China}\\[-.05cm]
  \texttt{\{shchen, yxliu, whan, wnzhang, tliu\}@ir.hit.edu.cn   }\\
}

\DeclareUnicodeCharacter{1F31F}{\textasteriskcentered} 

\begin{document}
\maketitle
\newcommand{\yxliu}[1]{\textcolor{orange}{#1}}
\renewcommand{\thefootnote}{\fnsymbol{footnote}}
\footnotetext[2]{Corresponding author.}

\renewcommand{\thefootnote}{\arabic{footnote}}

\newif\ifreview

\makeatletter
\@ifpackageloaded{acl}{
    \@ifclasswith{acl}{review}{
        \reviewtrue
    }{
        \reviewfalse
    }
}{
    \reviewfalse
}
\makeatother
\ifreview
    \newcommand{\repolink}{\url{https://anonymous.4open.science/r/Multi-Generative\_Agent\_System\_Survey/}}
\else
    \newcommand{\repolink}{\url{https://github.com/bianhua-12/Multi-generative_Agent_System_survey}}
\fi

\begin{abstract}
\Acf{LLM-MAS} have become a research hotspot since the rise of \acfp{LLM}. However, with the continuous influx of new related works, the existing reviews struggle to capture them comprehensively.
This paper presents a comprehensive survey of these studies. We first discuss the definition of \ac{LLM-MAS}, a framework encompassing much of previous work. We provide an overview of the various applications of \ac{LLM-MAS} in (i) solving complex tasks, (ii) simulating specific scenarios, and (iii) evaluating generative agents.
Building on previous studies, we also highlight several challenges and propose future directions for research in this field. 
\end{abstract}


\section{Introduction}
\begin{figure*}
    \centering
    \includegraphics[width=1\linewidth]{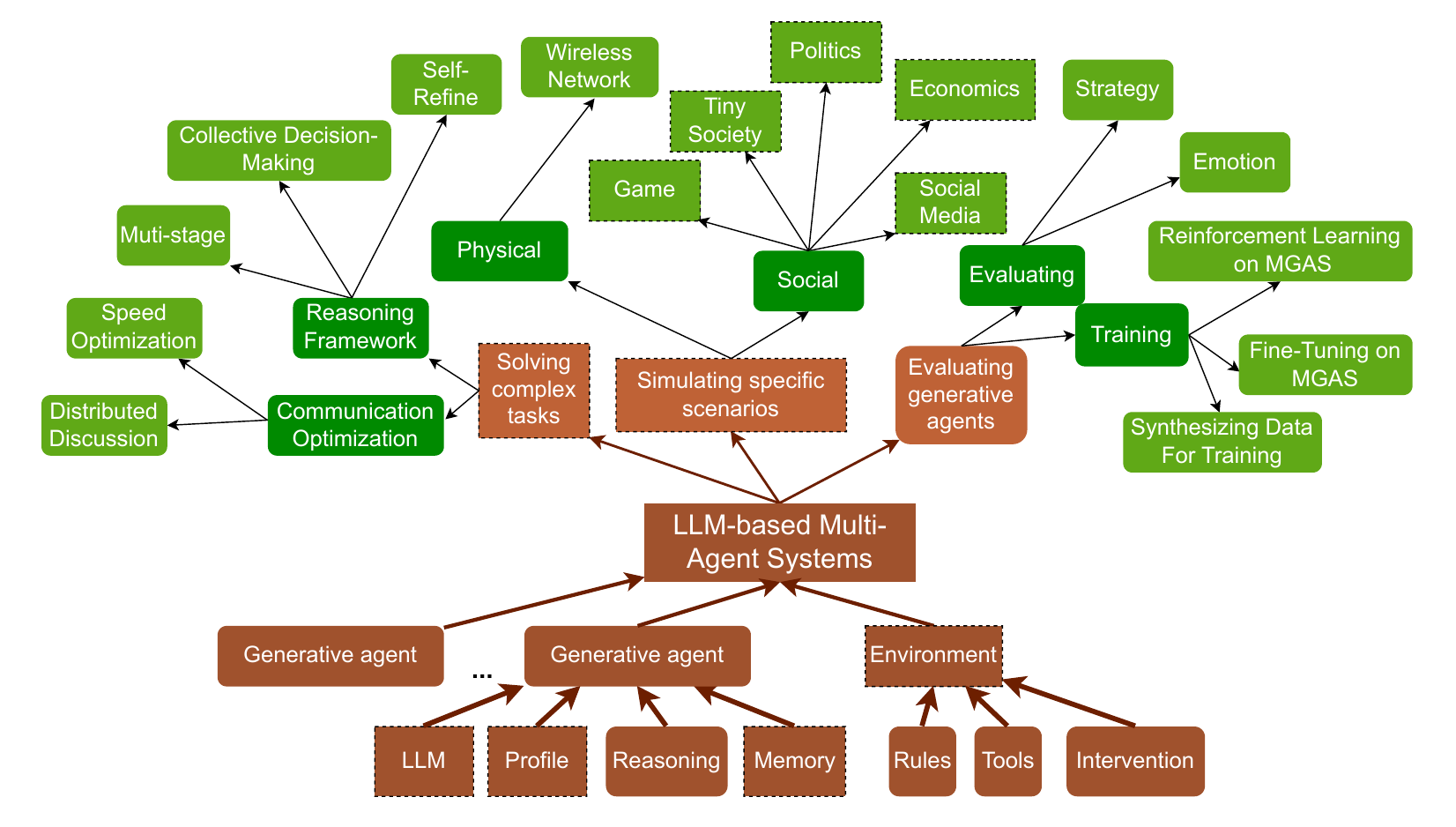}
    \caption{Overview of the application framework and relationship of LLM-MAS, generative agent, and LLM. Dashed-bordered right-angled rectangles represent content aligned with previous surveys, while rounded rectangles indicate original contributions introduced in this study. }
    \label{fig: application framework}
\end{figure*}

\Acf{MAS} have seen significant expansion owing to its adaptability and ability to address complex, distributed challenges~\cite{balajiIntroductionMultiAgentSystems2010}. 
Compared to single-agent settings~\cite{gronauerMultiagentDeepReinforcement2022}, \ac{MAS} provide a more accurate representation of the real world, as many real-world applications naturally involve multiple decision-makers interacting simultaneously. 
However, constrained by traditional~\acf{RL} agent parameters and the absence of general knowledge and capabilities, agents are unable to tackle complex decision-making tasks, such as collaborating with other agents for the development~\cite{qianChatDevCommunicativeAgents2024}. 
In recent years, \acfp{LLM}, e.g. Llama 3~\cite{dubeyLlama3Herd2024}, and GPT-4~\cite{openaiGPT4TechnicalReport2024}, have achieved notable successes, training on a massive web corpus~\cite{radfordLanguageModelsAre}. 
Compared with \ac{RL}, generative agents, with \ac{LLM} as the core control agents, can be better at reasoning, long-trajectory decision-making, etc., even without training~\cite{shinnReflexionLanguageAgents2023}.
Furthermore, generative agents offer natural language interfaces for interacting with humans, making these interactions more flexible and easier to explain~\cite{parkGenerativeAgentsInteractive2023}.
Based on these advantages, \acf{LLM-MAS} emerged. Researchers have surveyed these emerging works and proposed a general framework~\cite{guoLargeLanguageModel2024}. However, as the number of related studies continues to grow, some works have emerged that fall outside the scope of the original framework. 
In this paper, we provide a new perspective based on previous reviews of \acf{LLM-MAS} with a focus on recent advancements and discuss potential research directions. 
We collected 125 papers published in top artificial intelligence conferences, such as \textasteriskcentered ACL, NeurIPS, AAAI, and ICLR, in 2023 and 2024, along with some unpublished yet valuable papers from arXiv.\footnote{The list of papers included in this survey can be found in \repolink{}}
Based on the purpose of \ac{LLM-MAS}, we summarize the application of LLM-MAS as task-solving, simulation for specific problems, and evaluation of generative agents. Figure \ref{fig: application framework} illustrates the framework we propose for \ac{LLM-MAS} application. {(i)} Solving complex tasks\label{app1}. Multi-agents will naturally split tasks into subtasks, which will improve task performance. {(ii)} Simulating for specific scenarios\label{app2}. Researchers see \ac{LLM-MAS} as a sandbox for simulating problems in a specific domain. {(iii)} Evaluating generative agents\label{app3}. Compared with traditional task evaluation, \ac{LLM-MAS} has the capability of dynamic assessment, which is more flexible and harder for data leakage. For each category, we will discuss representative \ac{LLM-MAS}, resources, and their evaluation.

Compared to the previous survey~\cite{guoLargeLanguageModel2024,liSurveyLLMbasedMultiagent2024,hanLLMMultiAgentSystems2024,gronauerMultiagentDeepReinforcement2022}, this survey has the following distinctive contributions: (i) \textbf{A Taxonomy focusing on application of \ac{LLM-MAS}}: we introduce a more recent  taxonomy (taxonomy and difference are shown in Figure \ref{fig: application framework}) based on the purpose of the application of \ac{LLM-MAS}. (ii) \textbf{More Resources}: we analyze open-source frameworks and research works with benchmarks or datasets to facilitate the research community. (iii) \textbf{Challenges and Future}: we discuss the challenges in \ac{LLM-MAS}, and shed light on future research.
\section{Core Components of LLM-MAS}
\label{sec:LLM-MAS_definitions}

\ac{LLM-MAS} refer to a system that includes 
a collection of generative agents capable of interacting and collaborating within a shared environmental setting~\cite{wangRethinkingBoundsLLM2024}. We will discuss generative agents and the environment in the following.
\subsection{Generative Agents}

Generative agents refer to the components of \ac{LLM-MAS} that have role definitions, can perceive the environment, make decisions, and perform complex actions to change the environment~\cite{wangSurveyLargeLanguage2024}. They can be a player in a game or a user on social media and have the role of driving the development of \ac{LLM-MAS} and influencing its results.

Compared to traditional agents, generative agents need to be able to perform more complex behaviors, such as generating complete personalized blog posts based on historical information~\cite{parkSocialSimulacraCreating2022}. Therefore, in addition to using \acp{LLM} as the core, generative agents also require the following characteristics:
\begin{enumerate*}[label=(\roman*)]
    \item \emph{Profiling} is used to link their behavior by describing roles in natural language~\cite{gaoS3SocialnetworkSimulation2023}, or customizing the prompts for each generative agent based on their tasks~\cite{xuLanguageAgentsReinforcement2023}. 
    \item \emph{Memory} is used to store historical trajectories and retrieve relevant memories for subsequent agent actions, enabling the ability to take long-term actions while solving the problem of limited LLM context windows. There usually include three layers of memory: long-term, short-term, and sensory memory~\cite{parkGenerativeAgentsInteractive2023}.  
    \item \emph{Planning} is to formulate general behavior for a longer period of time in the future~\cite{yaoReActSynergizingReasoning2023}.
    \item \emph{Action} executes the interaction between the generative agent and the environment~\cite{wangSurveyLargeLanguage2024}. Generative agents may be required to choose one of several candidate behaviors to execute, such as voting for whom~\cite{xuExploringLargeLanguage2024}, or generate behaviors without mandatory constraints, such as generating a paragraph of text~\cite{liMetaAgentsSimulatingInteractions2023}. 
\end{enumerate*}

Generative agents can communicate with each other to achieve cooperation within the system.
The communication of generative agents can be roughly divided into two purposes. 
\begin{enumerate*}[label=(\roman*)]
    \item  The first purpose is to achieve collaboration, share the information obtained by themselves with other intelligent agents, and to some extent, aggregate multiple intelligent agents into a complete system, achieving performance beyond independent intelligent agents~\cite{yuanSkillReinforcementLearning2023};
    \item The second purpose is to achieve consensus, allowing for greater similarity in behavior or strategy among some agents, thereby enabling faster convergence to Nash equilibrium~\cite{oroojlooy2023review}.
\end{enumerate*}
 
The type of communication content can be roughly divided into two types: natural language and custom content. Natural language forms of communication have high interpretability and flexibility. Still, they are difficult to optimize, making them more suitable for pursuing consensus, such as Chatdev \cite{qianChatDevCommunicativeAgents2024} and job fair systems \cite{liMetaAgentsSimulatingInteractions2023}.  Custom content may be a vector or a discrete signal that no one can understand except for the generative agent in the system. But this form is easy to optimize using policy gradients, so it is commonly used for achieving cooperative purposes, such as the DIAL \cite {2015Deep} algorithm and its variables.
\subsection{Environment}

Environmental settings include rules, tools, and intervention interfaces:
\begin{enumerate*}[label=(\roman*)]
    \item \emph{Tools} are responsible for translating the agent’s action instruction into specific outcomes. Generative agents send action instructions to the environment and the environment converts the instruction into a record that the action was taken. There are different action spaces in different scenes. In the social media scene, the action space concludes ``like'', ``comment'', ``follow'', etc.~\cite{wangUserBehaviorSimulation2024}. 
    In the development scene, the action space closes the chat chain~\cite{qianChatDevCommunicativeAgents2024}, which is larger than social networks. 
    \item \emph{Rules} define the mode of communication between generative agents or the interaction with the environment, directly defining the behavioral structure of the entire system. Based on the scene, there are some special rules for the system, such as rules of the game~\cite{xuExploringLargeLanguage2024,chenPutYourMoney2024} and the norm of social behavior~\cite{parkGenerativeAgentsInteractive2023, wangUserBehaviorSimulation2024}. Normally, a generative agent in the large-scale system has a smaller action space and is more easily replaced by a rule-based model~\cite{mouUnveilingTruthFacilitating2024}.
    \item \emph{Intervention} provides an interface for external intervention systems. This intervention can come from any external source, human \cite{wangUserBehaviorSimulation2024}, or a supervision model \cite{chenPutYourMoney2024}, even a generative agent \cite{qianChatDevCommunicativeAgents2024}. The purpose of an intervention may be to actively read information from the system~\cite{wangUserBehaviorSimulation2024}, or passively interrupt the system to prevent uncontrolled behavior from occurring~\cite{qianChatDevCommunicativeAgents2024}.
\end{enumerate*}




\section{LLM-MAS for Solving Complex Tasks}
\label{sec:Collaboration learning}
\label{sec:LLM-MAS_for_tasks}

Completing a complex task usually requires multiple roles, multiple steps, and so on. This is difficult for a single agent, but multiple agents working together will be well suited to this task~\cite{islamMapCoderMultiAgentCode2024}. Further, each of these agents can be trained independently~\cite{shenSmallLLMsAre2024, yuNeekoLeveragingDynamic2024}. Compared with a single agent, \ac{LLM-MAS} can achieve better results. That is, the multi-agent collaboration will improve the overall performance \cite{duImprovingFactualityReasoning2023}. 


\subsection{Representative LLM-MAS for Solving Complex Tasks}
This field is currently a hot research topic. Recently, researchers mainly focus on multi-agent reasoning frameworks and multi-agent communication optimization, which will be discussed below.

\textbf{LLM-MAS reasoning framework.}
We summarize three aspects by the pipeline of reasoning, including: (i) multi-stage framework, (ii) collective decision-making framework, and (iii) self-refine framework. That is, the multi-stage framework refers to a pipeline where agents act as serial problem solvers at different stages~\cite{qianChatDevCommunicativeAgents2024}, while collective decision-making~\cite{zhaoElectoralApproachDiversify2024} refers to different agents voting or debating for one goal. Self-Refine refers to the mechanism of self-reflection in \ac{LLM-MAS}. Researchers propose a framework for applying multi-agents to the natural sciences~\cite{chenAutoAgentsFrameworkAutomatic2024a} to enhance data analysis, model simulations, and decision-making processes \cite{yinPEARRobustFlexible2024}. \citet{zhangProAgentBuildingProactive2023} propose a framework to achieve self-adaptation and adaptive cooperation. Scaling law in agent cooperation is also explored~\cite{qianScalingLargeLanguageModelbasedMultiAgent2024}, finding that there is no significant pattern.

\textbf{LLM-MAS communication optimization.}
The fully connected communication in \ac{LLM-MAS} can lead to issues such as combinatorial explosion and privacy disclosure. Based on this, we summarize two aspects in Communication Optimization, including: (i) speed optimization and (ii) distributed discussion. Speed optimization refers to researchers trying to speed up the communication of agents, for example, with non-verbal communication~\cite{liuDroidSpeakEnhancingCrossLLM2024} or shorter generation~\cite{chenOptimaOptimizingEffectiveness2024}. While distributed discussion refers to agents trying to solve tasks without enough information~\cite{liuAutonomousAgentsCollaborative2024}. Agents need to communicate with each other to achieve their goals~\cite{zhangProAgentBuildingProactive2023}, even without complete information in one agent\cite{liuAutonomousAgentsCollaborative2024}.

\subsection{Resources of LLM-MAS for Solving Complex Tasks}
We summarize common and open-source LLM-MAS for simulation in Table \ref{tab: collaboration}, including code, dataset, and benchmark. \\
\begin{table*}[ht]
    \centering
        \caption{Codes and Benchmarks in LLM-MAS for solving tasks studies. ``No Code'' or ``No Benchmark'' means the code or benchmark is unavailable.}
    \label{tab: collaboration}

\resizebox{0.92\textwidth}{!}{

    \begin{tabular}{ccccc} \toprule
   
 Field & SubField & Paper & Code& Dataset and Benchmark\\
\midrule 
\multirow{18}{2.3cm}{\makecell[c]{Reasoning \\Framework}}
&\multirow{7}{*}{Muti-stage}
&\cite{qianChatDevCommunicativeAgents2024}&  \href{https://github.com/OpenBMB/ChatDev}{Code Link} &SRDD\\
&&\cite{duMultiAgentSoftwareDevelopment2024}&  \href{https://github.com/OpenBMB/ChatDev}{Code Link} &SRDD\\
&&\cite{yueSynergisticMultiAgentFramework2024}&\href{https://github.com/yueshengbin/SMART}{Code Link}&SMART (self) \\
&&\cite{liuBOLAABenchmarkingOrchestrating2023}& \href{https://github.com/salesforce/BOLAA}{Code Link}&WebShop\\ 
&&\cite{linMAOFrameworkProcess2024}&\href{https://anonymous.4open.science/r/MAO-1074}{Code Link}&FG-C, CG-O \\
&&\cite{islamMapCoderMultiAgentCode2024}&\href{https://github.com/Md-Ashraful-Pramanik/MapCoder}{Code Link}&\makecell[c]{HumanEval, EvalPlus, MBPP, \\APPS, xCodeEval, CodeContest}\\
&&\cite{shenSmallLLMsAre2024}&\href{https://github.com/X-PLUG/Multi-LLM-Agent}{Code Link}&ToolBench, ToolAlpaca \\
\cmidrule(lr){2-5}
&\multirow{6}{*}{\makecell[c]{Collective \\Decision-Making}}
&\cite{zhaoElectoralApproachDiversify2024}&\href{https://github. com/xiutian/GEDI}{Code Link}&MCQA \\
&&\cite{chengCooperCoordinatingSpecialized2024}&\href{https://github.com/YiCheng98/Cooper}{Code Link}&ESConv dataset, P4G dataset \\
&&\cite{liangEncouragingDivergentThinking2024}&\href{https://github.com/Skytliang/Multi-Agents-Debate}{Code Link} & MT-Bench \\
&&\cite{leiMACMUtilizingMultiAgent2024}&\href{https://github.com/bin123apple/MACM}{Code Link}&MATH \\
&&\cite{zhangExploringCollaborationMechanisms2024a}&\href{https://github.com/zjunlp/MachineSoM}{Code Link}&MMLU, MATH, Chess Move Validity \\
&&\cite{wangUnleashingEmergentCognitive2024}&\href{https://github.com/MikeWangWZHL/Solo-Performance-Prompting.git}{Code Link}&TriviaQA \\
\cmidrule(lr){2-5}
&\multirow{5}{*}{\makecell[c]{Self-Refine}}
&\cite{wangRethinkingBoundsLLM2024}&\href{https://github.com/HKUST-KnowComp/LLM-discussion}{Code Link}&FOLIO-wiki \\
&&\cite{chenReConcileTableConference2024}&\href{https://github.com/dinobby/ReConcile}{Code Link}&\makecell[c]{StrategyQA, CSQA, GSM8K, AQuA, \\MATH, Date Understanding, ANLI}\\
&&\cite{chenAutoAgentsFrameworkAutomatic2024a}&\href{https://github.com/Link-AGI/AutoAgents}{Code Link}&TriviaQA \\
&&\cite{tangCodeAgentAutonomousCommunicative2024}&\href{https://github.com/Code4Agent/codeagent}{Code Link}&Trans-Review,AutoTransform,T5-Review \\
&&\cite{zhangProAgentBuildingProactive2023}&\href{https://pku-proagent.github.io/}{Code Link}&Overcooked-AI\\
\midrule
\multirow{3}{2.3cm}{\makecell[c]{Communication \\ Optimization}}
&\multirow{1}{*}{Speed Optimization}
&\cite{liuDroidSpeakEnhancingCrossLLM2024}&No Code&HotpotQA,NarrativeQA,MultifieldQA\\
\cmidrule(lr){2-5}
&\multirow{2}{*}{Distributed}
&\cite{chenInternetAgentsWeaving2024}&\href{https://github.com/OpenBMB/IoA}{Code Link}&\makecell[c]{TriviaQA, Natural Questions, \\HotpotQA, 2WikiMultiHopQA} \\
&&\cite{liuAutonomousAgentsCollaborative2024}&\href{https://github.com/thinkwee/iAgents}{Code Link}&InformativeBench\\
        \bottomrule
    \end{tabular}
}
\end{table*}
\textbf{Data set.} All datasets of traditional NLP tasks are available. In addition, following ECL \cite{qianExperientialCoLearningSoftwareDeveloping2024}, \citet{qianChatDevCommunicativeAgents2024} evaluate the quality of generated software on the SRDD dataset and systematically evaluate agent capabilities in the domain of software development.

\textbf{Open source community.} The open-source and industrial communities have also contributed significantly to the development of LLM-MAS. MetaGPT \cite{hongMetaGPTMetaProgramming2023} assigns different roles to generative agents to form a collaborative entity for complex tasks. \citet{gaoAgentScopeFlexibleRobust2024} propose AgentScope with message exchange as its core communication mechanism. In the meantime, this work develops a distribution framework that facilitates seamless switching between local and distributed deployments and automatic parallel optimization with minimal effort.
Open AI {proposes} Swarm~\cite{OpenaiSwarm2024}, an experimental multi-agent orchestration framework that is ergonomic and lightweight. Unlike the previously released Assistants API, Swarm gives developers fine-grained control over context, steps, and tool calls rather than being hosted. 

\subsection{Evaluation of LLM-MAS for solving complex task}
\textbf{Performance on specific tasks.}
Shown as Table \ref{tab: collaboration}, the performance of LLM-MAS can be evaluated by specific tasks, which is intuitive and convenient. For example, in an APP system \cite{zhangAppAgentMultimodalAgents2023}, the average number of steps and tools used by an agent to complete a specific task are considered as indicators; in BOLAA \cite{liuBOLAABenchmarkingOrchestrating2023}, the recall and QA accuracy of intelligent physical examination retrieval are also considered as evaluation indicators; in the Werewolf game \cite{xuLanguageAgentsReinforcement2023}, the win rate of virtual players is naturally also an evaluation indicator; in the job fair system \cite{liMetaAgentsSimulatingInteractions2023} , the proportion of correctly recruited target job seekers by the recruiting party is also an evaluation indicator; in the auction system \cite{chenPutYourMoney2024}, the Spearman correlation coefficient between the predicted and actual prices of goods, as well as the skills of bidders, are also measured by TrueSkill scores \cite{graepel2007bayesian}; in Stanford Town \cite{parkGenerativeAgentsInteractive2023}, the quality of behaviors generated by virtual agents and human agents is manually sorted and evaluated using TrueSkill; in urban simulation systems \cite{xuUrbanGenerativeIntelligence2023}, the success rate of completing specific tasks such as navigation is also an evaluation metric.

\textbf{Communication cost analysis.}
The paramount concern lies in the operational cost of the system. Given that a substantial proportion of contemporary systems incorporate \acp{LLM} as a pivotal module, the additional expenditure incurred during system operation has emerged as a pivotal area of interest. As an illustrative example,  \citet{mouUnveilingTruthFacilitating2024} utilize the actual runtime of the system as a pivotal metric, underscoring the significance of managing this operational overhead. 
\section{LLM-MAS for Simulating Specific Scenarios}
\label{sec:LLM-MAS_for_simulation}

This section will illustrate the application for \ac{LLM-MAS} in simulation. Researchers apply agents to simulate a certain scenario to study its impact on a specific subject like social science. On the one hand, compared with rule-based methods~\cite{chuangComputationalAgentbasedModels2023a}, generative agents with natural language communication can be more intuitive for humans. On the other hand, environment determines the properties of the simulation, which is the core of the entire simulation.
\subsection{Representative LLM-MAS for Simulating Specific Scenarios}
The typical scenarios for \ac{LLM-MAS} simulations are described as follows. We will introduce the following work according to the subject. 

\textbf{Social domain.}
Social large-scale experiments in the real world have high costs, and the sheer scale of social participation can sometimes escalate into violence and destruction, posing potential ramifications~\cite{mouUnveilingTruthFacilitating2024}. Therefore, it is necessary to simulate in the virtual environment; simulation can solve the problem of excessive overhead in the real environment and can simulate the process in the real world for a long time at a faster speed \cite{liAgentHospitalSimulacrum2024}. At the same time, the whole process can be easily repeated, which is conducive to further research. 
Researchers have done a lot of work to simulate social media scenarios. Based on the social media simulation archetype \cite{parkSocialSimulacraCreating2022}, \citet{parkGenerativeAgentsInteractive2023} propose Stanford Town, which leads to a one-day simulation of the life of 25 agents with different occupations in a small American town. At the same time, there was work on emotional propagation influence \cite{gaoS3SocialnetworkSimulation2023}, information cocoon room based on recommendation scenario research \cite{wangUserBehaviorSimulation2024}, and study of social movements \cite{mouUnveilingTruthFacilitating2024}. Researchers propose Urban Generative Intelligence (UGI) \cite{xuUrbanGenerativeIntelligence2023} to address specific urban issues and simulate complex urban systems, providing a multidisciplinary approach to understanding and managing urban complexity. \citet{liAgentHospitalSimulacrum2024} study doctor agent evolution method by hospital simulation. Because doctor agent training is both inexpensive and highly effective, this work can quickly scale up the agent to handle tens of thousands of cases in just a few days, a task that would take a human doctor years to complete.~\citet{panVeryLargeScaleMultiAgent2024} propose a huge scale of agent simulation, increasing the number of agents to $10^6$.
In social game,like Werewolf~\cite{xuExploringLargeLanguage2024} , Avalon~\cite{lanLLMBasedAgentSociety2024} , and Minecraft \cite{gongMindAgentEmergentGaming2024} for \ac{LLM-MAS} simulation are attempted. {Further, some game companies like Netease are also actively experimenting with \ac{LLM-MAS} in their games}.

\textbf{Physical domain.}
For the physical domain, the applications for generative agent simulation include mobility behaviors, transportation~\cite{gaoLargeLanguageModels2023}, wireless networks, etc. However, there is limited research in the area of multi-generative agents. \citet{zouWirelessMultiAgentGenerative2023} explore the application of multiple agents in the wireless field, proposing a framework where multiple on-device agents can interact with the environment and exchange knowledge to solve a complex task together.
\subsection{Resources for LLM-MAS simulation}

We summarize common and open-source \ac{LLM-MAS} for simulation in Table \ref{tab: sim}, including code and benchmarks. 

To prove the effectiveness of the simulation, that is, to fit the reality, researchers usually evaluate the simulation system by simulating real data. Therefore, a realistic dataset with dense users and records is very important for evaluation simulation~\cite{mouUnveilingTruthFacilitating2024}. An ideal dataset will be dense: that is, data with a smaller number of users on the same scale can better evaluate the simulation capability of the \ac{LLM-MAS}.

For Benchmark, \citet{duHelmsmanMassesEvaluate2024} propose WWQA based on werewolf scenarios to evaluate the agent's capability in a werewolf scenario. SoMoSiMu-Bench \cite{mouUnveilingTruthFacilitating2024} provides evaluation benchmarks focusing on individual user behavior and social simulation system results.

\begin{table*}[!t]
\centering
\caption{{Codes and Benchmarks in LLM-MAS for simulation studies. ``No Code'' or ``No Benchmark'' means the code or benchmark is unavailable.}}
\label{tab: sim}
\resizebox{0.8\textwidth}{!}{
\begin{tabular}{ccccc}\toprule
Domain & Subdomain & Paper & Code & Dataset and Benchmark\\
\midrule
\multirow{11}{*}{\makecell[c]{Social}}
& \multirow{5}{*}{Tiny Society}
&\cite{huangAdaSocietyAdaptiveEnvironment2024} & No Code & AdaSociety\\
&&\cite{chenAgentCourtSimulatingCourt2024} & \href{https://github.com/relic-yuexi/AgentCourt}{Code Link} & AgentCourt  \\
&&\cite{parkGenerativeAgentsInteractive2023} & \href{https://github.com/joonspk-research/generative_agents}{Code Link} & No Benchmark or Dataset\\
&&\cite{piattiCooperateCollapseEmergence2024} & \href{https://github.com/giorgiopiatti/govsim}{Code Link} & No Benchmark\\
&&\cite{chuangSimulatingOpinionDynamics2024} & \href{https://github.com/yunshiuan/llm-agent-opinion-dynamics}{Code Link} & No Benchmark or Dataset\\
\cmidrule(lr){2-5}
& \multirow{1}{*}{Economics}
&\cite{liEconAgentLargeLanguage2024} & \href{https://github.com/tsinghua-fib-lab/ACL24-EconAgent}{Code Link} &  No Benchmark or Dataset\\
\cmidrule(lr){2-5}
& \multirow{3}{*}{Social Media}
&\cite{wangUserBehaviorSimulation2024} & \href{https://github.com/RUC-GSAI/YuLan-Rec}{Code Link} & Movielens-1M\\
&&\cite{gaoS3SocialnetworkSimulation2023} & No Code & Blog Authorship Corpus\\
&&\cite{mouUnveilingTruthFacilitating2024} & \href{https://github.com/xymou/social_simulation}{Code Link} & SoMoSiMu-Bench(self) \\
\cmidrule(lr){2-5}
& \multirow{2}{*}{Game}
&\cite{duHelmsmanMassesEvaluate2024} & \href{https://github.com/doslim/Evaluate-the-Opinion-Leadership-of-LLMs}{Code Link} & WWQA\\
&&\cite{panVeryLargeScaleMultiAgent2024} & \href{https://github.com/modelscope/agentscope/tree/main/examples/paper_large_scale_simulation}{Code Link} & No Benchmark or Dataset\\

\midrule
\multirow{1}{*}{Physical}
& Wireless&\cite{zouWirelessMultiAgentGenerative2023}  & No Code & No Benchmark or Dataset\\

    \bottomrule
\end{tabular}
}
\end{table*}

\subsection{Evaluation of LLM-MAS simulation} 
\label{sec: Simulation Evaluation}
We will discuss the evaluation based on indicators used for assessing LLM-MAS as a whole, rather than the capabilities of individual agents.

\textbf{Consistency.}
\ac{LLM-MAS} necessitate a robust congruence with the real world to ensure the derivation of meaningful and insightful experimental outcomes. In the context of simulation systems, exemplified by UGI \cite{xuUrbanGenerativeIntelligence2023}, the primary objective lies in faithfully replicating specific real-world scenarios. When employed for training agents like SMART \cite{yueSynergisticMultiAgentFramework2024}, only those agents that have undergone rigorous training within a virtual environment that closely mirrors the real environment can be deemed suitable for deployment in real-world settings. Similarly, when utilized for evaluation purposes, such as in AgentSims \cite{linAgentSimsOpenSourceSandbox2023}, the attainment of authentic and reliable evaluation results is contingent upon the virtual environment maintaining a high degree of consistency with its real-world counterpart. Finally, in the system for collecting data such as BOLAA \cite{liuBOLAABenchmarkingOrchestrating2023}, consistency also ensures the validity of the data. Therefore, an important performance measure of LLM-MAS is its consistency with the real situation. 

 

\textbf{Information dissemination.}
Compare the differences between information dissemination behavior in the system and reality using time series analysis methods. Information dissemination can to some extent reflect the nature of media; therefore, a realistic multi-agent system should have a similar information dissemination trend to the real world. \citet{abdelzaher2020multiscale} compare the changes in the number of events occurring each day in an online social media simulation environment; S3 \cite{gaoS3SocialnetworkSimulation2023} compare the number of users who are aware of a certain event every day, as well as the changes in emotional density and support rate for that event every day; a similar approach is also used in Stanford Town \cite{parkGenerativeAgentsInteractive2023}. 


\section{LLM-MAS for Evaluating Generative Agents}
\label{sec:LLM-MAS_for_evaluation}

With \acp{LLM} prevailing in the community, how to evaluate the ability of \acp{LLM} is an open question. Existing evaluation methods suffer from the following shortcomings: {(i)} constrained evaluation abilities, {(ii)} vulnerable benchmarks, and {(iii)} unobjective metrics. The complexity and diversity of \ac{LLM-MAS} have indicated that \ac{LLM-MAS} can evaluate \acp{LLM}. However, how to design specific evaluation indicators and evaluation methods has puzzled researchers. 
Similarly, \ac{LLM-MAS} can also be used in training generative agents. We summarize three aspects of training: {(i) Supervised Fine-Tuning (SFT) (ii) \acf{RL} (iii) Synthesizing data for training}.
\subsection{Representative LLM-MAS for Evaluating Generative Agents} 
\ac{LLM-MAS} can provide rewards to agents, and these rewards can be used to evaluate or train generative agents, which will be discussed below. 

\textbf{Evaluation of generative agents.}
\label{subsection:evaluation on LLMs}
Researchers study generative agents by putting them into \ac{LLM-MAS}. In \ac{LLM-MAS}, researchers can further study the LLM's strategic capabilities in different scenes, such as long strategic ability \cite{chenPutYourMoney2024}, leadership strategy~\cite{xuLanguageAgentsReinforcement2023} and competitiveness strategy~\cite{zhaoCompeteAIUnderstandingCompetition2024}. In the emotional field, MuMA-ToM \cite{shiMuMAToMMultimodalMultiAgent2024} is used to evaluate the ability of agents to understand and reason about human interactions in a real home environment through video and text descriptions.

\textbf{Training on generative agents.} \citet{liCoEvolConstructingBetter2024} enhance the data to { Supervised Fine-Tuning (SFT)} generative agents with \ac{LLM-MAS}.~\citet{xuLanguageAgentsReinforcement2023} have created generative agents to overcome the intrinsic bias from \acp{LLM} by proposing a novel framework that powers generative agents with multi-agent reinforcement learning \cite{xuLanguageAgentsReinforcement2023}. For \ac{LLM-MAS}, \citet{yueSynergisticMultiAgentFramework2024} split complex trajectories in knowledge-intensive tasks into subtasks, proposing a co-training paradigm of the multi-agent framework, Long- and Short-Trajectory Learning, which ensures synergy while keeping the fine-grained performance of each agent. RLHF has been criticized for its high cost. \citet{liuTrainingSociallyAligned2023} propose an alignment scheme based on a multi-agent system, effectively addressing instability and reward gaming concerns associated with reward-based \ac{RL} optimization. Either way, \ac{LLM-MAS} are essentially viewed as an environment in \ac{RL} with different ways of getting rewards from the environment.

\subsection{Resources of LLM-MAS for evaluations}
Table \ref{tab: eval} shows the work with code, dataset, and benchmark we summarize, serving as a reference for future researchers. 
\begin{table*}[ttt]
\centering
\caption{{Codes and Benchmarks in LLM-MAS for evaluation studies. ``No Code'' or ``No Benchmark'' means the code or benchmark is unavailable.}}
\label{tab: eval}

\resizebox{0.92\textwidth}{!}{
\begin{tabular}{ccccc}\toprule
Domain & Subdomain & Paper & Code& Dataset and Benchmark\\

\midrule
\multirow{11}{*}{\makecell[c]{Evaluation of \\ generative agents}}
& \multirow{7}{*}{Strategy}
&\cite{liuAgentBenchEvaluatingLLMs2023} & \href{https://github.com/THUDM/AgentBench}{Code Link}& AGENTBENCH\\
&&\cite{bandiAdversarialMultiAgentEvaluation2024} & No Code & MT-Bench\\
&&\cite{chanChatEvalBetterLLMbased2023a}&\href{https://github.com/thunlp/ChatEval}{Code Link} & ChatEval\\
&&\cite{chenLLMArenaAssessingCapabilities2024}&\href{https://github.com/THU-BPM/LLMArena.}{Code Link}& LLMARENA\\
&&\cite{xuMAgICInvestigationLarge2023}&\href{https://github.com/cathyxl/MAgIC}{Code Link}&MAgIC\\
&&\cite{huangMLAgentBenchEvaluatingLanguage2024}&\href{https://github.com/snap-stanford/MLAgentBench}{Code Link}&MLAgentBench\\
&&\cite{chenPutYourMoney2024}&\href{https://github.com/jiangjiechen/auction-arena}{Code Link}&AUCARENA\\

 \cmidrule(lr){2-5}
& \multirow{2}{*}{Emotion}
&\cite{zhangPsySafeComprehensiveFramework2024}&\href{https://github.com/AI4Good24/PsySafe}{Code Link}&PsySafe\\
&&\cite{shiMuMAToMMultimodalMultiAgent2024}&\href{https://github.com/SCAI-JHU/MuMMA-ToM}{Code Link}&MuMA-ToM\\
\midrule
\multirow{4}{*}{\makecell[c]{Training on \\ generative agents}}
& \multirow{1}{*}{SFT on LLM-MAS}
&\cite{liCoEvolConstructingBetter2024}&\href{https://github.com/lirenhao1997/CoEvol}{Code Link}&MT-Bench, AlpacaEval \\
\cmidrule(lr){2-5}
& \multirow{1}{*}{MARL on LLM-MAS}
&\cite{xuLanguageAgentsReinforcement2023}& No Code & No dataset or benchmark\\
\cmidrule(lr){2-5}
& \multirow{1}{*}{Synthesized Ddata}
&\cite{liuTrainingSociallyAligned2023}&\href{https://github.com/agi-templar/Stable-Alignment}{Code Link}&\makecell[c]{HH, Moral Stories, MIC, \\ETHICS-Deontology, TruthfulQA}\\
    \bottomrule
\end{tabular}
}
\end{table*}

\section{Challenges and Future Directions }
\label{sec:LLM-MAS_challenges}

While previous work on \ac{LLM-MAS} has obtained many remarkable successes, this field is still at its initial stage, and there are several significant challenges that need to be addressed in its development. In the following, we outline several key challenges along with potential future directions.

\subsection{Challenges posed by generative agents}
Generative agents are an integral part of \ac{LLM-MAS}. However, the generative agents have some shortcomings due to the inherent characteristics of the base model \acp{LLM}, which will be carefully discussed below.

\textbf{Challenges. }
{(i)} Generalized alignment for simulation~\cite{liuTrainingSociallyAligned2023}. When the agents are leveraged for real-world simulation, a perfect generative agent should be able to depict diverse traits~\cite{wangSurveyLargeLanguage2024} honestly. However, due to the training method of the foundation model~\cite{openaiGPT4TechnicalReport2024}, generative agents usually cannot be aligned with mock objects.  {(ii)} Hallucination. Generative agents have a certain probability of hallucination in their interaction with other agents~\cite{duImprovingFactualityReasoning2023}. Various enhancement methods can alleviate this problem but cannot solve it {(iii)} Lack of sufficient long text capability. When processing complex information, generative agents forget the input information because of the lack of long-text ability~\cite{zhaoLongAgentScalingLanguage2024}.

\textbf{Future directions.} The improvement of the ability of a single agent or the ability of the base model has always been a hot topic. Researchers have focused on enhancing alignment, reducing hallucination, and improving the ability of long text. The proposal of the new generation of Open AI  model o1 \cite{IntroducingOpenAIO12024}, provides researchers with new ideas, that is, to use \textbf{more complex reasoning} to enhance the ability of the model.

\subsection{Challenges posed by interactions}
Due to the complexity, autoregressive, and other characteristics of \ac{LLM-MAS}, there are many problems in the practical application of the system. Two main problems, \emph{Efficiency explosion}, and \emph{Accumulative Effect}, are listed in the following.

\textbf{Efficiency explosion.}
Due to their autoregressive architecture, \acp{LLM} typically have slow inference speeds. However, generative agents need to query \acp{LLM} for each action multiple times, such as extracting information from memory, making plans before taking actions, and so on. When the \ac{LLM-MAS} scales up, this problem will be magnified, especially for generative agents that have a large action space. SoMoSiMu-Bench~\cite{mouUnveilingTruthFacilitating2024} replaces the edge generative agents with rule-based agents, alleviating this problem. However, for \ac{LLM-MAS} with a complex action space in generative agents, this problem remains unsolved. 

\textbf{Accumulative Effect.}
Since each round of \ac{LLM-MAS} is based on the results of the previous round, and \ac{LLM-MAS} have a great impact on the subsequent results. Researchers have used a rule-based model for intermediate error correction \cite{chenPutYourMoney2024}, but there is still a lot of room for improvement. IOA \cite{chenInternetAgentsWeaving2024} proposed an Internet-like communication architecture, which made LLM-MAS more scalable and enhanced the adaptability to dynamic tasks.

\textbf{Future directions.} Industry academia has been making efforts to reduce the communication cost of \ac{LLM-MAS}, such as alignment-based method OPTIMA \cite{chenOptimaOptimizingEffectiveness2024} and Industrialized parallel message method AgentScope \cite{gaoAgentScopeFlexibleRobust2024}, but it is still in the basic stage and has a large research space.
\subsection{Challenges of Evaluating for LLM-MAS}
\textbf{Lack of Objective metrics for group behavior.}
As shown in Section \ref{sec: Simulation Evaluation}, due to the diversity, complexity, and unpredictability of multi-agent environments, it is difficult to obtain sufficiently detailed, specific, and direct system evaluation indicators from current work at the population level. At present, researchers mainly compare the distribution of the system and real environments to evaluate \ac{LLM-MAS}, which lacks details for the \ac{LLM-MAS} running process.

\textbf{Automated evaluation and benchmark.} Different \ac{LLM-MAS} of the same kind cannot be compared because of the lack of a benchmark for \ac{LLM-MAS}. Further, there is a lack of a common benchmark framework for both individual and total-based evaluation,  that can be used to evaluate most \ac{LLM-MAS}.

\textbf{Future directions.} Studying large-scale \ac{LLM-MAS} will be a new research hotspot, from which researchers will evaluate and discover new scale effects. In the meantime, common test benchmarks and evaluation methods will also emerge in future research.

\section{Conclusion}
In this survey, we systematically summarize existing research in the \acf{LLM-MAS} field. We present and review these studies from three 
application aspects: task-solving, simulation, and evaluation of the \ac{LLM-MAS}. We provide a detailed taxonomy to draw connections among the existing research, summarizing the major techniques and their development histories for each of these aspects. In addition to reviewing the previous work, we also propose several challenges in this field, which are expected to guide potential future directions.
\section*{Limitations}
We have made our best effort, but some limitations may still exist. Due to page limitations, we can only provide a brief summary of each method without exhaustive technical details. On the other hand, we primarily collect studies from \textasteriskcentered ACL, NeurIPS, ICLR, AAAI, and arXiv, and there is a chance that we may have missed some important work published in other venues. In application, we primarily list representative \ac{LLM-MAS} resources with open code in Table \ref{tab: collaboration}, Table \ref{tab: sim}, and Table  \ref{tab: eval}. More complete papers can be found in \repolink{}. We recognize the timeliness of our work, and we will stay abreast of discussions within the research community, updating opinions and supplementing overlooked work in the future.

\section*{Acknowledgments}
This research was supported by the National Key Research and Development Program (No. 2022YFF0902100), and the Nature Scientific Foundation of Heilongjiang Province (YQ2021F006).

\bibliography{latex/generative_agent,latex/multi_generative_agent_sim, latex/hw}

\appendix




\end{document}